\documentclass{article}
\usepackage[numbers]{natbib}
\usepackage{auxhook}

\usepackage[preprint]{neurips_2024}

\usepackage[utf8]{inputenc} 
\usepackage[T1]{fontenc}    
\usepackage{url}            
\usepackage{booktabs}       
\usepackage{amsfonts}       
\usepackage{nicefrac}       
\usepackage{microtype}      
\usepackage{graphicx} 
\usepackage[table]{xcolor}
\usepackage{xcolor}   
\usepackage[utf8]{inputenc}
\usepackage{booktabs} 
\usepackage{arabtex} 
\usepackage{utf8} 
\usepackage{parskip}
\usepackage{subcaption} 
\usepackage{multirow}
\usepackage[unicode,pdfencoding=auto]{hyperref}

\hypersetup{
    colorlinks=true,
    linkcolor=violet,
    urlcolor=violet,
    citecolor=violet
}

\setcode{utf8}
\renewcommand{\arraystretch}{1.5}
\newcommand{\total}[1]{\cellcolor{gray!20}\textcolor{black!}{#1}} 
\newcommand{\best}[1]{\textcolor{green!50!black}{#1}}
 
\newcommand{\besty}[1]{\cellcolor{green!10}\textcolor{green!50!black}{#1}}
\newcommand{\second}[1]{\underline{#1}} 

\pdfobj{1 0 R}
\title{%
  \centering
  \begin{minipage}[c]{0.10\textwidth}
    \centering
    \includegraphics[height=1.7cm]{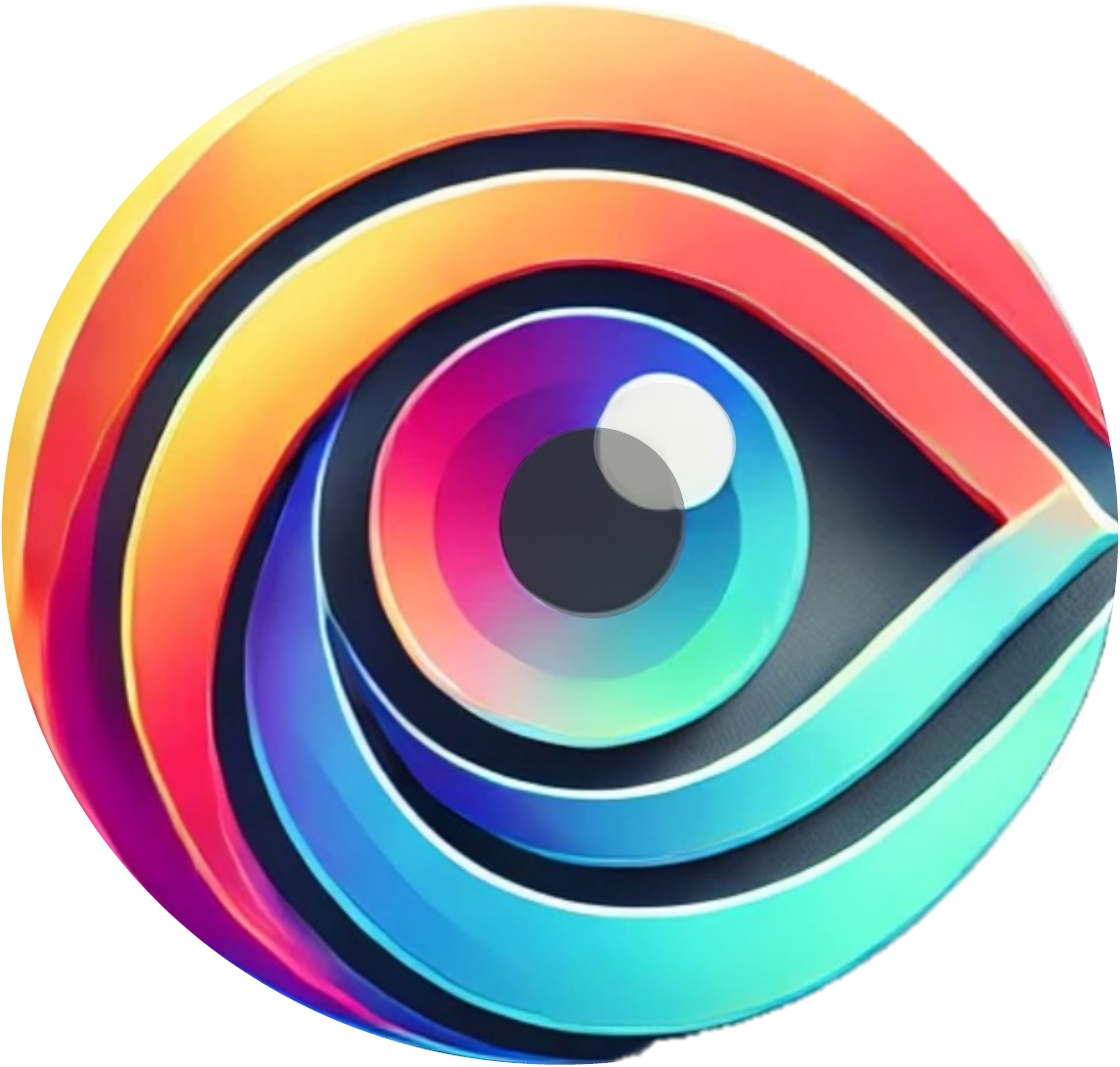}
  \end{minipage}%
  \hspace{1.15em}%
  \begin{minipage}[c]{0.6\textwidth}
    \centering 
    \textbf{\LARGE AIN: The Arabic INclusive Large Multimodal Model}
  \end{minipage}
}

\author{
    Ahmed Heakl\textsuperscript{1}\thanks{Equal Contribution} \quad
    Sara Ghaboura\textsuperscript{1*}  \quad
    Omkar Thawakar\textsuperscript{1} \\
    \begin{tabular}{@{}c@{}}
        \textbf{Fahad Shahbaz Khan\textsuperscript{1,2} \hspace{0.2em}
        Hisham Cholakkal\textsuperscript{1} \hspace{0.2em}
        Rao Muhammad Anwer\textsuperscript{1,3} \hspace{0.2em}
        Salman Khan\textsuperscript{1,4}}
    \end{tabular} \\[0.7em]
    \textsuperscript{1}Mohamed bin Zayed University of AI, 
    \textsuperscript{2}Linköping University, 
    \textsuperscript{3}Aalto University, \\
    \textsuperscript{4}Australian National University
}

\begin{document}
\maketitle

\begin{center}
    \begin{tabular}{@{}l l@{\hspace{1cm}} l l@{\hspace{1cm}} l l@{}}
        $\vcenter{\hbox{\includegraphics[height=0.65cm]{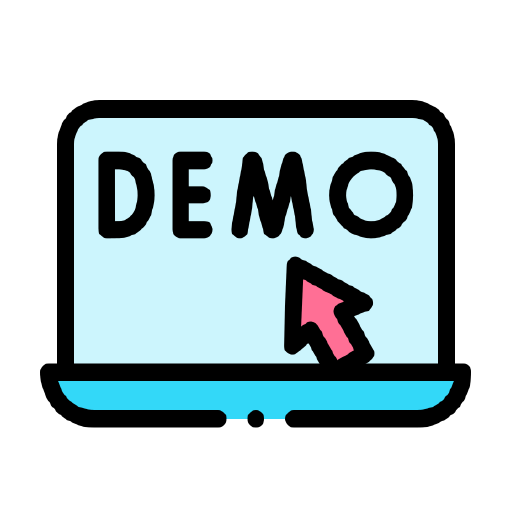}}}$ &
        $\vcenter{\hbox{\href{https://huggingface.co/spaces/ahmedheakl/AIN-Arabic-VLM}{\texttt{AIN Demo}}}}$ &
        $\vcenter{\hbox{\includegraphics[height=0.55cm]{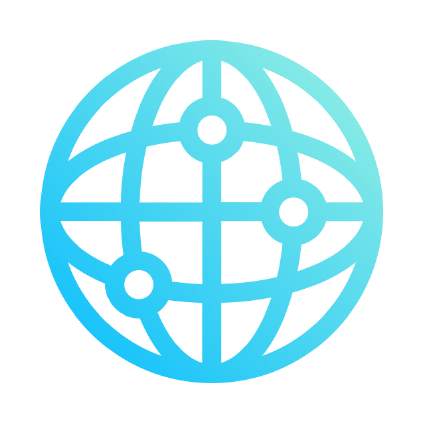}}}$ &
        $\vcenter{\hbox{\href{https://mbzuai-oryx.github.io/AIN/}{\texttt{AIN Webpage}}}}$ &
        $\vcenter{\hbox{\includegraphics[height=0.55cm]{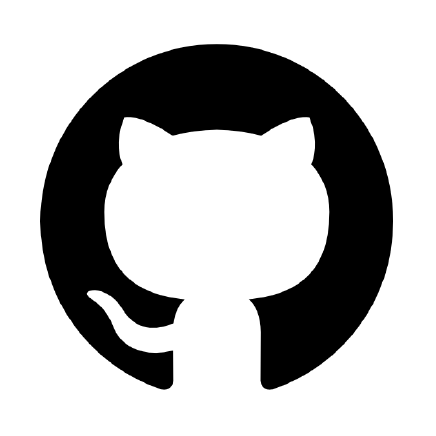}}}$ &
        $\vcenter{\hbox{\href{https://github.com/mbzuai-oryx/AIN}{\texttt{AIN GitHub}}}}$ 
    \end{tabular}
\end{center}

\begin{figure}[ht]
\centering
  \includegraphics[width=12cm,height=3.5cm]{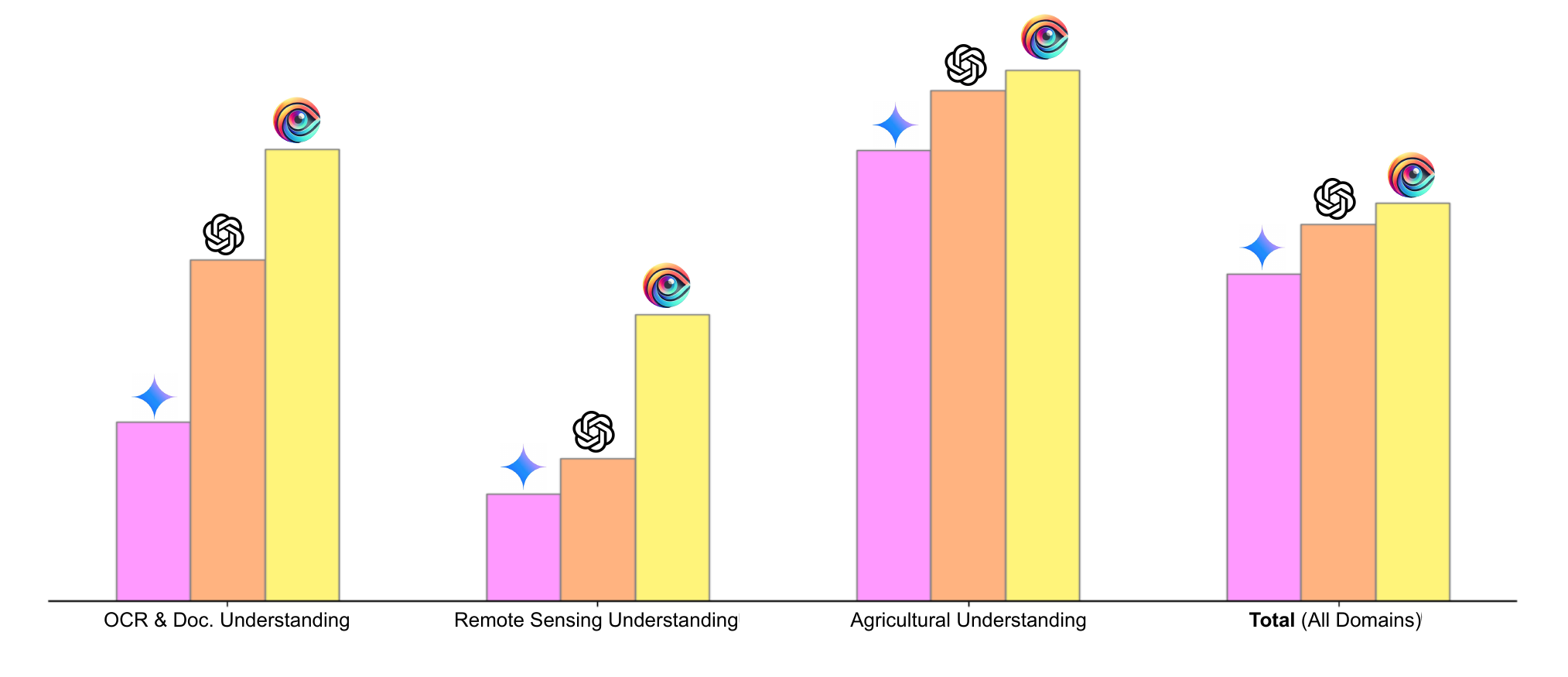} 
  \caption{\small{Cross-domain performance analysis on the Camel-Bench Benchmark.
  Our AIN-7B achieves promising performance compared to significantly bigger models (GPT-4o and Gemini-1.5-Pro)
   in both domain-specific and aggregate settings. Despite its smaller size, our AIN-7B achieves competitive performance across all 38 sub-domains with significantly superior capabilities on OCR \& document understanding.}}
  \label{fig:radar_models_perf}
\end{figure}

\begin{abstract}
Amid the swift progress of large language models (LLMs) and their evolution into large multimodal models (LMMs), significant strides have been made in high-resource languages such as English and Chinese. While Arabic LLMs have seen notable progress, Arabic LMMs remain largely unexplored, often narrowly focusing on a few specific aspects of the language and visual understanding. To bridge this gap, we introduce \emph{AIN—}the \emph{Arabic Inclusive Multimodal Model}—designed to excel across diverse domains. AIN is an English-Arabic bilingual LMM designed to excel in English and Arabic, leveraging carefully constructed 3.6 million high-quality Arabic-English multimodal data samples. AIN demonstrates state-of-the-art Arabic performance, while also possessing strong English-language visual capabilities. On the recent CAMEL-Bench benchmark comprising 38 sub-domains including, multi-image understanding, complex visual perception, handwritten document understanding, video understanding, medical imaging, plant diseases, and remote sensing-based land use understanding, our AIN demonstrates strong performance with the 7B model outperforming GPT-4o by an absolute gain of 3.4\% averaged over eight domains and 38 sub-domains. AIN's superior capabilities position it as a significant step toward empowering Arabic speakers with advanced multimodal generative AI tools across diverse applications. 
\end{abstract}

\section{AIN Capabilities}
\begin{figure}[t]
  \centering
  \includegraphics[width=14.5cm ,height=8cm ]{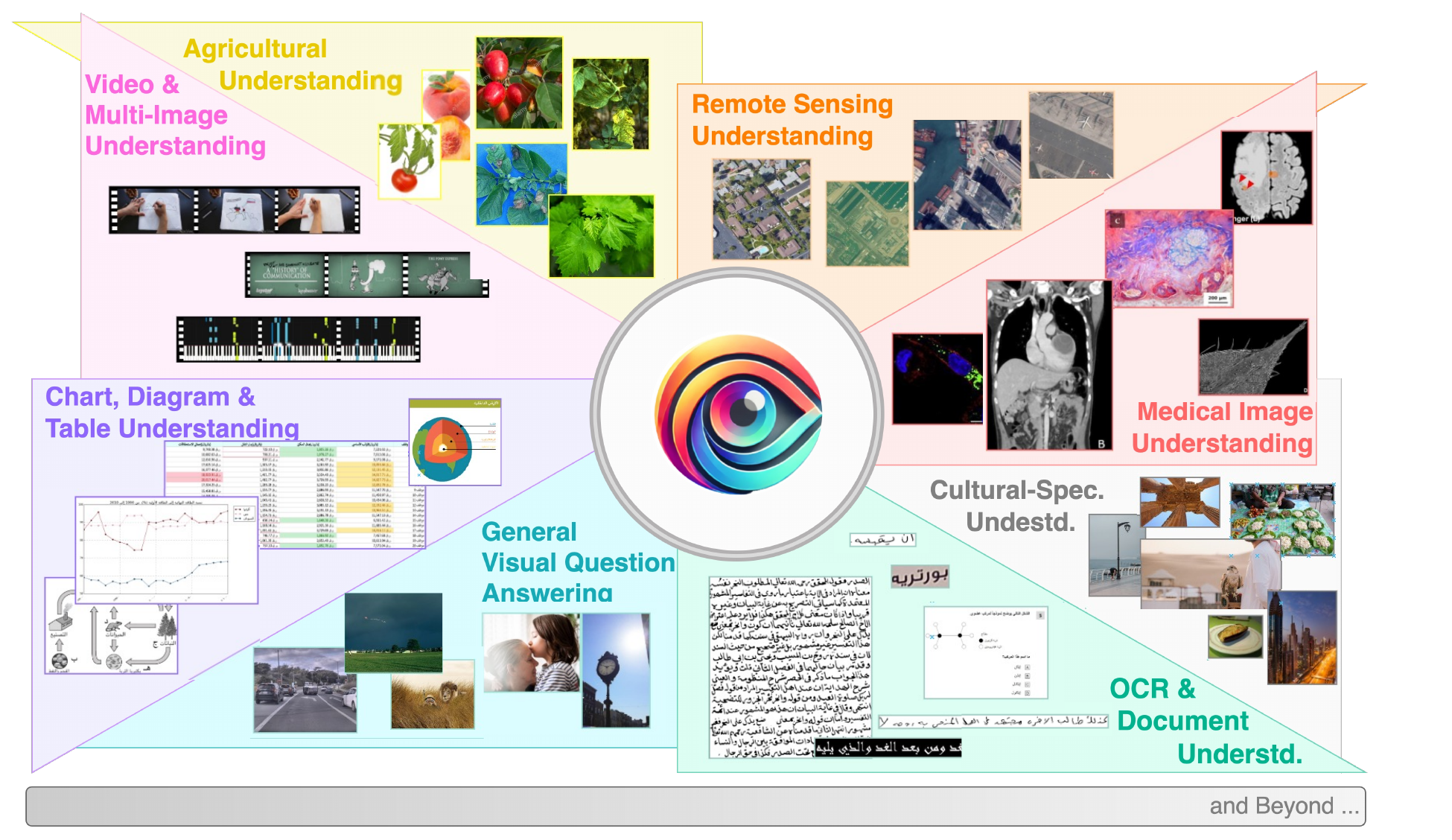} 
  \caption{\small{AIN: A versatile LMM excelling in visual and contextual understanding across diverse domains, including VQA on complex topics, OCR for various fonts and handwriting, cultural insights (traditions, food, places), agricultural tasks (crop identification, fruit classification, disease detection), remote sensing (multi-scale objects), medical imaging (various modalities), and video analysis (animation, human activities).}}
  \label{fig:ain_can_see}
\end{figure}

\begin{figure}[hptb]
  \centering
  \includegraphics[width=0.8\textwidth ]{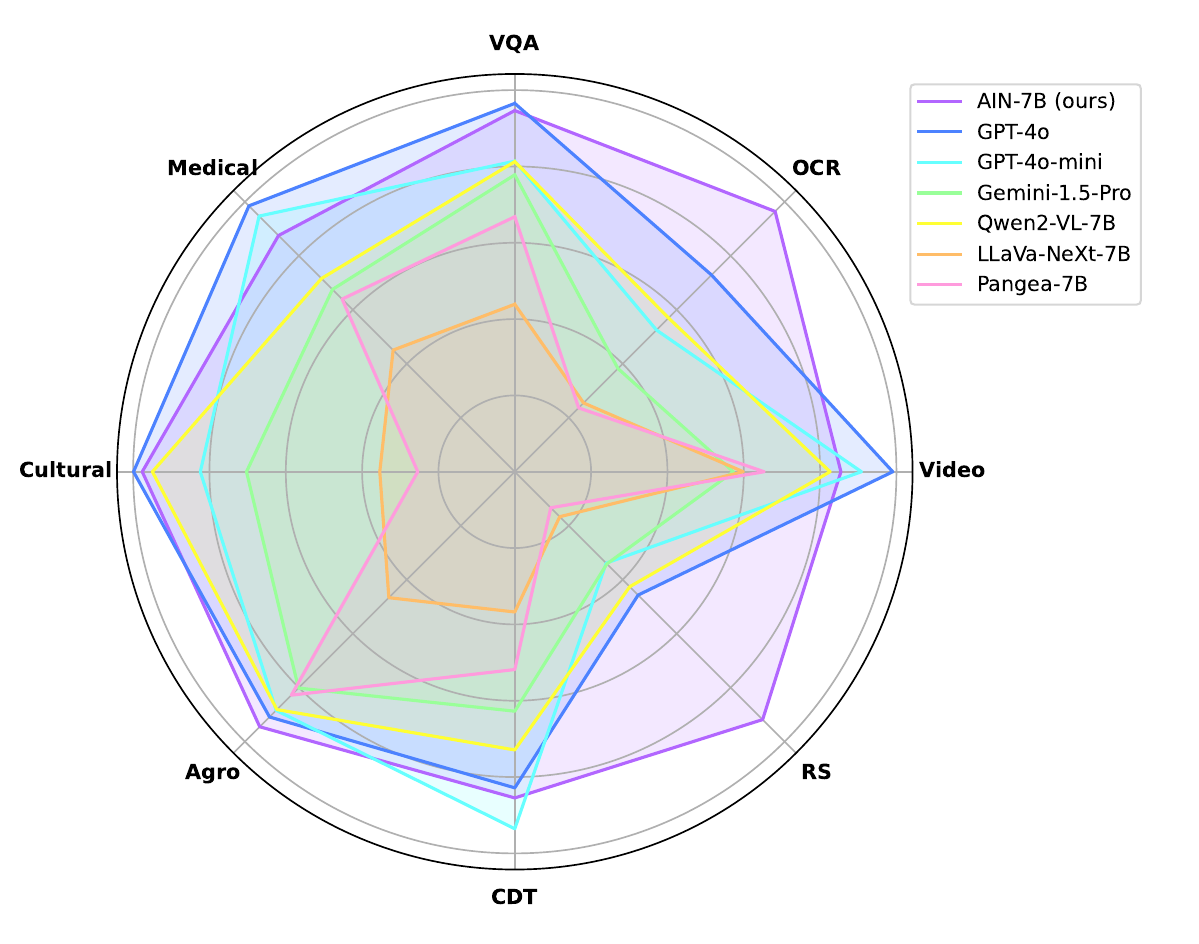} 
  \caption{AIN compared to existing LMMs across CAMEL-Bench benchmark \cite{ghaboura2024camel} domains: \small{\textbf{OCR}: ``OCR \& Document Understanding'', \textbf{Video}: ``General Video \& Multi-Image Understanding'', \textbf{RS}: ``Remote Sensing Understanding'', \textbf{CDT}:``Chart, Diagram \& Table Understanding'', \textbf{Agro.}: ``Agricultural Image Understanding'', \textbf{Cultural}: ``Cultural-Specific Understanding'',  \textbf{Medical}: ``Medical Image Understanding''.}}
  \label{fig:ain_radar}
\end{figure}

The AIN model is an advanced Arabic Large Multimodal Model (See Figure~\ref{fig:ain_can_see}) with strong English proficiency, built on 7 billion parameters derived from the Qwen-2-VL-7B architecture \cite{bai2023qwen}. Its performance highlights significant progress in multimodal understanding, excelling in complex reasoning, cross-lingual tasks, and detailed image-text alignment across diverse benchmarks.

\subsection{Quantitative Results}
We present the quantitative performance of the model across benchmarks and domains.
Figure~\ref{fig:ain_radar} showcases a comprehensive performance analysis of AIN-7B across CAMEL-Bench \cite{ghaboura2024camel} domains, comparing it with a variety of models including, GPT-4o \cite{gpt4o}, GPT-4o-mini \cite{gpt4o-mini}, Gemini-1.5-Pro \cite{team2024gemini}, Qwen2-VL-7B \cite{bai2023qwen}, LlaVA-NeXt-7B \cite{liu2024llavanext}, and Pangea-7B \cite{yue2024pangea}.
The plot demonstrates the superior performance of AIN-7B in most domains.
Table~\ref{tab:Camel_perf} shows that our AIN-7B achieves the best overall performance across all domains, compared to existing models.

\begin{table}[hptb]
\centering
\caption{Performance comparison of AIN and different closed- and open-source LMMs across CAMEL-Bench domains \cite{ghaboura2024camel}. \besty{Best} performance is highlighted in green; \second{second-best} performance is underlined. \\ \small{ \textbf{OCR\textsuperscript{*}}: ``OCR \& Document Understanding'', \textbf{Video\textsuperscript{*}}: ``General Video \& Multi-Image Understanding'', \textbf{RS\textsuperscript{*}}: ``Remote Sensing Understanding'', \textbf{CDT\textsuperscript{*}}:``Chart, Diagram \& Table Understanding'', \textbf{Agro.\textsuperscript{*}}: ``Agricultural Image Understanding'', \textbf{Cult.\textsuperscript{*}}: ``Cultural-Specific Understanding'',  \textbf{Med.\textsuperscript{*}}: ``Medical Image Understanding''.}}
\vspace{0.5em}
\renewcommand{\arraystretch}{1.5}
\setlength{\tabcolsep}{9pt} 
\resizebox{\textwidth}{!}{ 

\begin{tabular}{lccccccccc}
\hline
\textbf{Models} & \textbf{VQA} & \textbf{OCR} & \textbf{Video} & \textbf{RS} & \textbf{CDT} & \textbf{Agro.} & \textbf{Cult.} & \textbf{Med.} & \textbf{\cellcolor{gray!20}{Total}} \\
\hline

\small{GPT-4o} \cite{gpt4o} & \second{55.15} & \second{54.98} & \besty{69.65} & 27.36 & \second{62.35} & \second{80.75} & \besty{80.86} & \besty{49.91} & \second{\cellcolor{gray!20}{60.13}} \\
\small{GPT-4o-mini} \cite{gpt4o-mini} & 48.83 & 39.38 & \second{66.28} & 16.93 & 56.37 & 78.80 & 65.92 & \second{47.37} & \cellcolor{gray!20}{52.49} \\
\small{Gemini-1.5-Pro} \cite{gemini_announcement} & 46.68 & 28.68 & 42.95 & 17.07 & 47.06 & 72.14 & 56.24 & 33.78 & \cellcolor{gray!20}{52.38} \\
\small{Gemini-1.5-flash} \cite{gemini_announcement} & 45.59 & 27.58 & 53.31 & 14.95 & 48.26 & 76.07 & 46.54 & 42.87 & \cellcolor{gray!20}{44.40} \\

\small{InternVL-8B} \cite{chen2023internvl} & 30.41 & 15.91 & 51.42 & 5.36 & 30.27 & 44.47 & 20.88 & 29.48 & \cellcolor{gray!20}{28.52} \\
\small{InternVL2.5-1B} \cite{chen2024expanding} & 27.22 & 19.45 & 38.20 & 3.39 & 30.75 & 39.53 & 35.68 & 21.27 & \cellcolor{gray!20}{26.94} \\
\small{Qwen-VL-2B} \cite{bai2023qwen} & 41.02 & 22.93 & 38.90 & 12.56 &27.83 & 52.02 & 34.28 & 29.12 & \cellcolor{gray!20}{32.33} \\
\small{Qwen2-VL-7B} \cite{bai2023qwen} & 48.76 & 42.73 & 61.97 & 21.30 &54.67 & 79.32 & 75.96 & 35.81 & \cellcolor{gray!20}{52.56} \\
\small{LLaVa-NeXt-7B} \cite{liu2024llavanext} & 26.33 & 19.12 & 44.90 &8.33 &27.56 & 42.00 & 28.30 & 22.54 & \cellcolor{gray!20}{27.39} \\
\small{LLaVa-OneVision} \cite{li2024llava} & 42.90 & 31.35 & 29.41 & 10.72 & 40.86 & 75.03 & 66.02 & 27.29 & \cellcolor{gray!20}{40.45} \\
\small{Pangea-7B} \cite{yue2024pangea} & 40.09 & 17.75 & 49.01 & 6.67 &38.87 & 74.51 & 20.34 & 31.99 & \cellcolor{gray!20}{34.90} \\
\small{Maya-8B} \cite{alam2024maya} & 39.07 & 26.70 & 47.23 & \second{27.53} & 34.25 & 70.61 & 57.42 & 31.57 & \cellcolor{gray!20}{41.80} \\

\hline
\small{\textbf{AIN-7B} \emph{(ours)} }& \besty{56.78} & \besty{72.35} & 64.09 & \besty{45.92} & \besty{64.10} & \besty{85.05} & \second{78.09} & 43.77 & \besty{63.77} \\
\hline

\end{tabular}}
\label{tab:Camel_perf}
\end{table}

We conduct a comprehensive evaluation of the bilingual capabilities of AIN in Arabic and English using well-established benchmarks. To further assess AIN’s proficiency in Arabic beyond CAMEL-Bench, we evaluate its performance across all topics of ArabicMMLU \cite{koto2024arabicmmlu}. As shown in Table~\ref{tab:mmlu_perf}, AIN outperforms in 14 out of 19 evaluated categories, achieving a significant 3\% overall improvement compared to Qwen2-VL-7B \cite{bai2023qwen}. Similarly, as presented in Table~\ref{tab:bench_eng_perf}, AIN demonstrates strong English language capabilities on ten benchmarks that cover general VQA, mathematics, science, and visual chart interpretation.

\begin{table}[hptb]
\caption{ArabicMMLU all categories' performance comparison across various models. \best{Best} performance is highlighted in green. \\ 
\small{\textbf{Comp. Sci\textsuperscript{*}}: Computer Science,\textbf{ Gen. Knldge\textsuperscript{*}}: General Knowledge, \textbf{Islamic Std.\textsuperscript{*}}: Islamic Studies, \textbf{Managmt}: Management, \textbf{Nat. Sci\textsuperscript{*}}: Natural Science, \textbf{Political Sci.\textsuperscript{*}}: Political Science, \textbf{Social Sci.\textsuperscript{*}}: Social Science.}}
\vspace{0.5em}
\centering
\renewcommand{\arraystretch}{1.3} 
\setlength{\tabcolsep}{9pt} 
\resizebox{\textwidth}{!}{
\begin{tabular}{lccccc}
\hline
\rowcolor{gray!10}\textbf{Model} & \textbf{Accounting} & \textbf{Arabic Lang.} & \textbf{Biology} & \textbf{Civics} & \textbf{Comp. Sci.\textsuperscript{*}} \\
\hline
Qwen2-VL-7B \cite{bai2023qwen}  & 52.7  & 51.34 & \best{44} & \best{48.92 }& 65.87 \\
\small{\textbf{AIN-7B} \emph{(ours)}} & \best{59.46} & \best{55.41} & 43.72 & 47.37 & \best{69} \\
\hline
&  &  & &  &  \\
\hline
\rowcolor{gray!10}\textbf{Model} & \textbf{Driving Test} & \textbf{Economics} & \textbf{Gen. Knldge\textsuperscript{*}} & \textbf{Geography} & \textbf{History} \\
\hline
Qwen2-VL-7B \cite{bai2023qwen}  & 67.88 & \best{59.59} & 55.18 & 49.67 & 43.1 \\
\textbf{AIN-7B} \emph{(ours)}& \best{69.69} & 59.25 & \best{57.01 }& \best{55.16} & \best{45.16} \\
\hline
&  &  & &  &  \\
\hline
\rowcolor{gray!10}\textbf{Model} & \textbf{Islamic Std.\textsuperscript{*}} & \textbf{Law} & \textbf{Managmt\textsuperscript{*}} & \textbf{Math} & \textbf{Nat. Sci.\textsuperscript{*}} \\
\hline
Qwen2-VL-7B \cite{bai2023qwen}  & 55.93 & 58.92 & \best{68} & 64.79 & 71.45 \\
\textbf{AIN-7B} \emph{(ours)} &\best{ 58.64} & \best{71.66} & 64 & \best{66.99 }& \best{78.37} \\
\hline
&  &  & &  &  \\
\hline
\rowcolor{gray!10}\textbf{Model} & \textbf{Philosophy} & \textbf{Physics} & \textbf{Political Sci.\textsuperscript{*}} & \textbf{Social Sci.\textsuperscript{*}} & \textbf{\total{Total}} \\
\hline
Qwen2-VL-7B \cite{bai2023qwen} & 53.85 & 39.61 & 54.29 & \best{65.75} & \total{56.36} \\
\textbf{AIN-7B} \emph{(ours)} & \best{56.41} & \best{45.1} & \best{60} & 65.43 & \total{\best{59.36}}\\
\hline
\end{tabular}
}
\label{tab:mmlu_perf}
\end{table}


\begin{table}[hptb]
\centering
\caption{Comprehensive performance comparison of AIN-7B against Qwen2-VL-7B across \textbf{10 English benchmarks}, with relative performance gain ($\uparrow$) indicated. \best{Best} performance is highlighted in green. Our AIN-7B achieves promising performance on English language across these diverse benchmarks.\\
\small{Benchmarks: MMBench \cite{MMBench}, MME \cite{fu2023mme}, MMMU \cite{yue2023mmmu}, POPE\cite{Li-hallucination-2023}, SEED\cite{li2024seed}, MathVista \cite{lu2023mathvista}, ScienceQA \cite{lu2022learn}, ChartQA \cite{masry2022chartqa}, AI2D \cite{kembhavi2016diagram}, MMT-Bench \cite{ying2024mmtbench}.}}
\vspace{0.5em}
\renewcommand{\arraystretch}{1.5}
\setlength{\tabcolsep}{12pt} 
\resizebox{\textwidth}{!}{
\begin{tabular}{l*{5}{c}}
\hline
\rowcolor{gray!10}\textbf{Model} & 
\textbf{MMBench} &
\textbf{MME} &
\textbf{MMMU}&
\textbf{POPE} &
\textbf{SEED} \\
\hline
Qwen2-VL-7B \cite{bai2023qwen} & 81.78 & 1,675.90 & 52 & 86.13 & 76.42\\
\small{\textbf{AIN-7B} \emph{(ours)}} & \best{93.76} & \best{1,689.02} & \best{54.2} & \best{87.59} & \best{78.35}  \\
\hline
\rowcolor{gray!20}\textbf{Improved by\quad\textcolor{green!50!black}{$\uparrow$}} & \textcolor{green!50!black}{$\uparrow$} 11.98 & \textcolor{green!50!black}{$\uparrow$} 13.12 & \textcolor{green!50!black}{$\uparrow$} 2.2 & \textcolor{green!50!black}{$\uparrow$} 1.46 & \textcolor{green!50!black}{$\uparrow$} 1.93 \\
\hline
 &  &  &  & & \\
\hline
\rowcolor{gray!10}\textbf{Model} & 
\textbf{MathVista} &
\textbf{ScienceQA} &
\textbf{ChartQA} &
\textbf{AI2D} &
\textbf{MMT-Bench}\\
\hline
Qwen2-VL-7B \cite{bai2023qwen} & 61.19 & 85.87 & 83.16 & 82.77 & 63.29 \\
\small{\textbf{AIN-7B} \emph{(ours)}}& \best{63.9} & \best{91.82} & \best{85.12} & \best{83.29} & \best{63.64} \\
\hline
\rowcolor{gray!20}\textbf{Improved by\quad\textcolor{green!50!black}{$\uparrow$}}& \textcolor{green!50!black}{$\uparrow$} 2.71& \textcolor{green!50!black}{$\uparrow$} 5.95 & \textcolor{green!50!black}{$\uparrow$} 1.96 & \textcolor{green!50!black}{$\uparrow$} 0.52 & \textcolor{green!50!black}{$\uparrow$} 0.35 \\

\hline
\end{tabular}
}
\label{tab:bench_eng_perf}
\end{table}



\subsection{Qualitative Results}
\begin{figure}[ht]
  \centering
  \includegraphics[width=\textwidth]{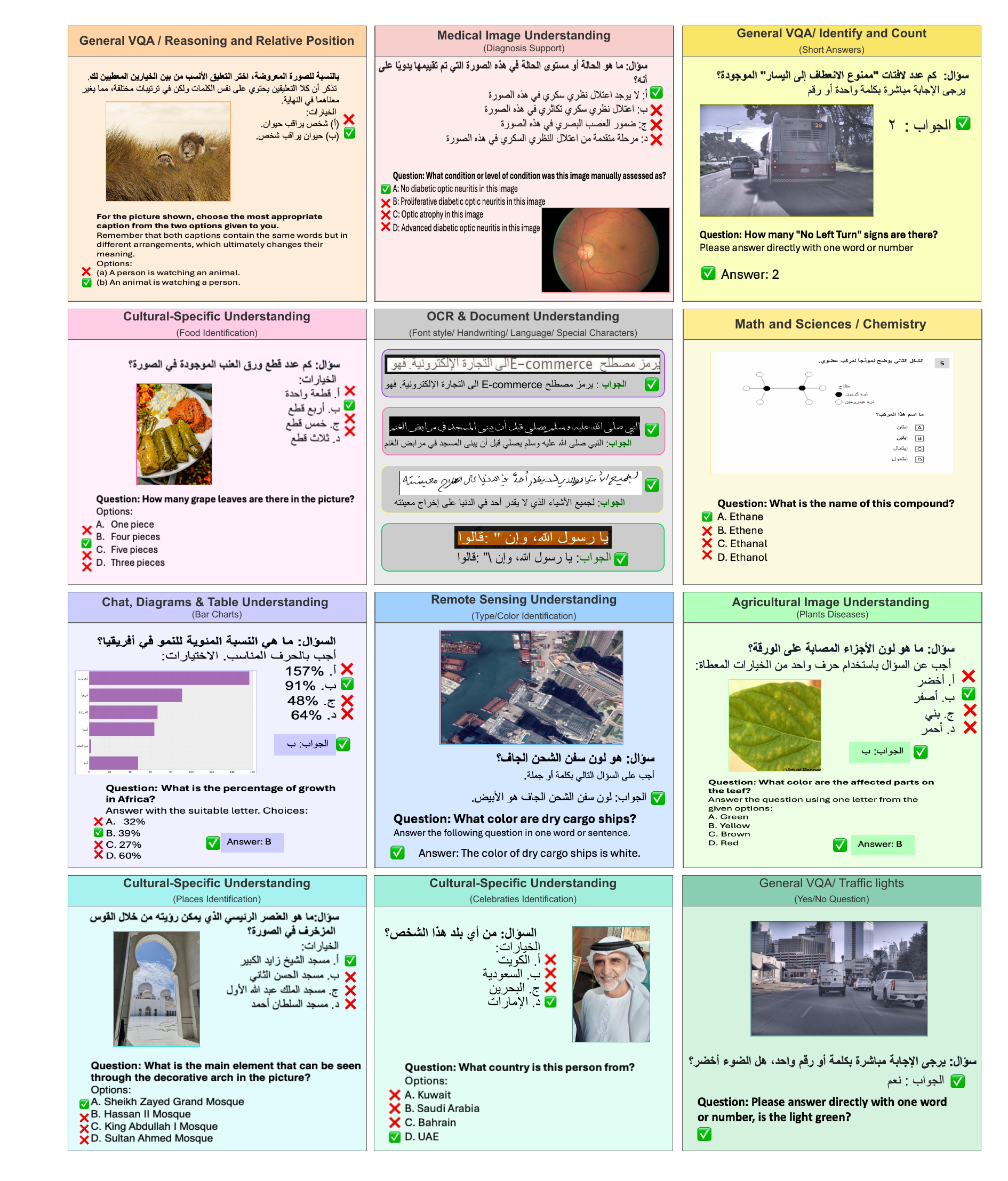}
  \caption{\small{Qualitative results demonstrating AIN's comprehensive capabilities across diverse domains. The results show its proficiency in handling both multiple-choice and open-ended questions. Our proposed AIN exhibits robust performance in addressing queries related to visual attributes (shape, color, quantity), while maintaining appropriate response formats (single character, word, or complete sentence) according to task requirements.}}
  \label{fig:ain_qual}
\end{figure}

The qualitative assessment of AIN's contextual response generation reveals its proficiency across diverse domains, as illustrated in Figure~\ref{fig:ain_qual}. The model excels in general VQA tasks, demonstrating advanced capabilities in OCR and document analysis, including both machine-printed text and handwriting interpretation. Additionally, it achieves prominent performance in medical image interpretation, scientific visualization comprehension, and remote sensing analysis. The model's capabilities extend to intricate data visualization interpretation, where it demonstrates skilled extrapolative understanding from charts and diagrams. Notably, AIN exhibits strong cultural-specific understanding through object recognition across diverse contexts, such as places, food items, and celebratory scenes. 
Throughout all tasks, AIN consistently produces accurate, contextually relevant, and comprehensive responses, underscoring its versatility in handling complex visual-linguistic tasks.


\begin{figure}
  \centering
  \includegraphics[width=0.95\textwidth]{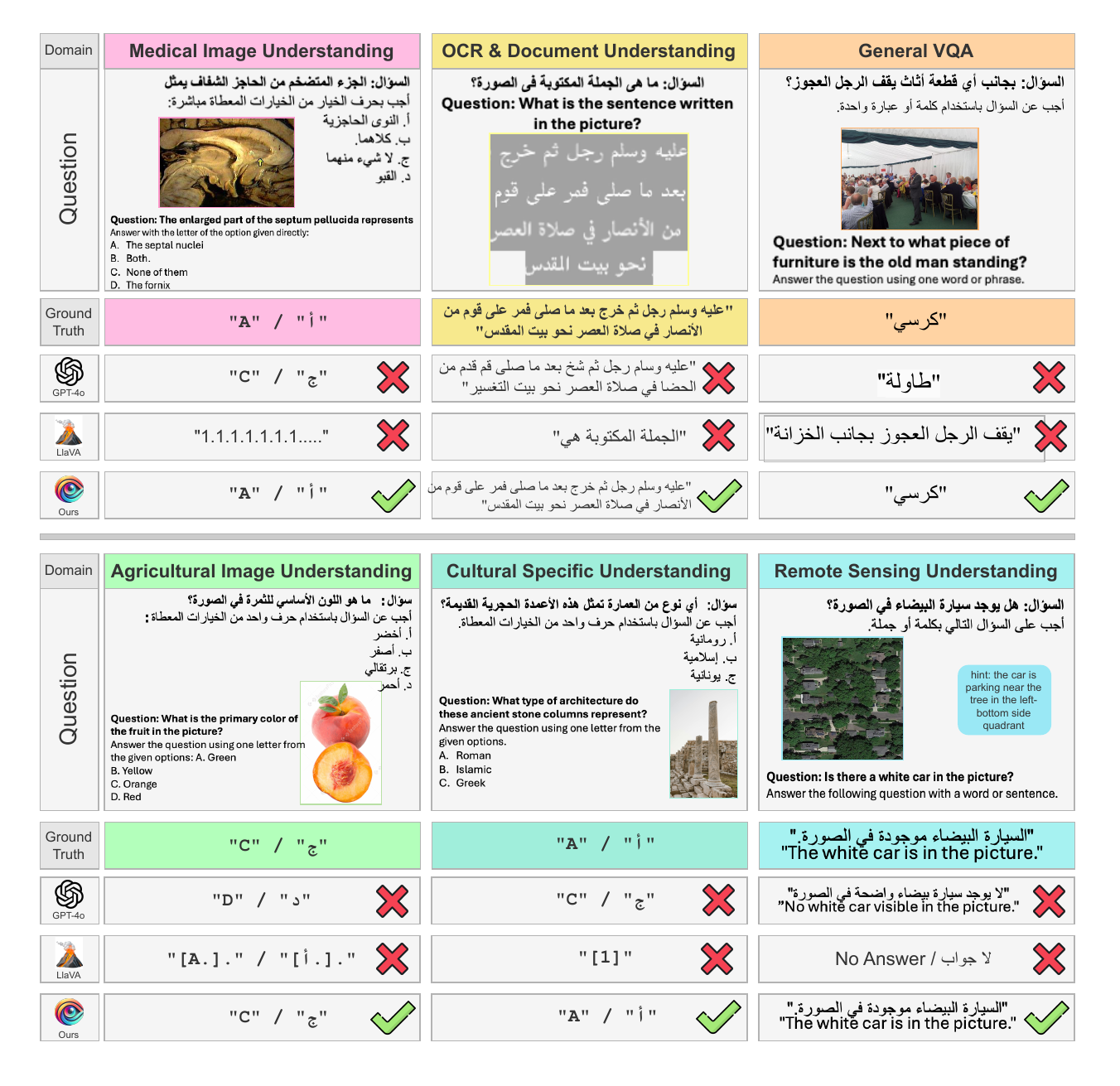} 
  \caption{Comparison of AIN with GPT-4o \cite{gpt4o} and LLaVA \cite{li2024llava}  across diverse tasks. The evaluation demonstrates AIN's proficiency in handling both multiple-choice and open-ended questions while maintaining appropriate response formats.}
  \label{fig:ain_contre}
\end{figure}

\begin{figure}[htpb]
  \centering
  \includegraphics[width=\textwidth]{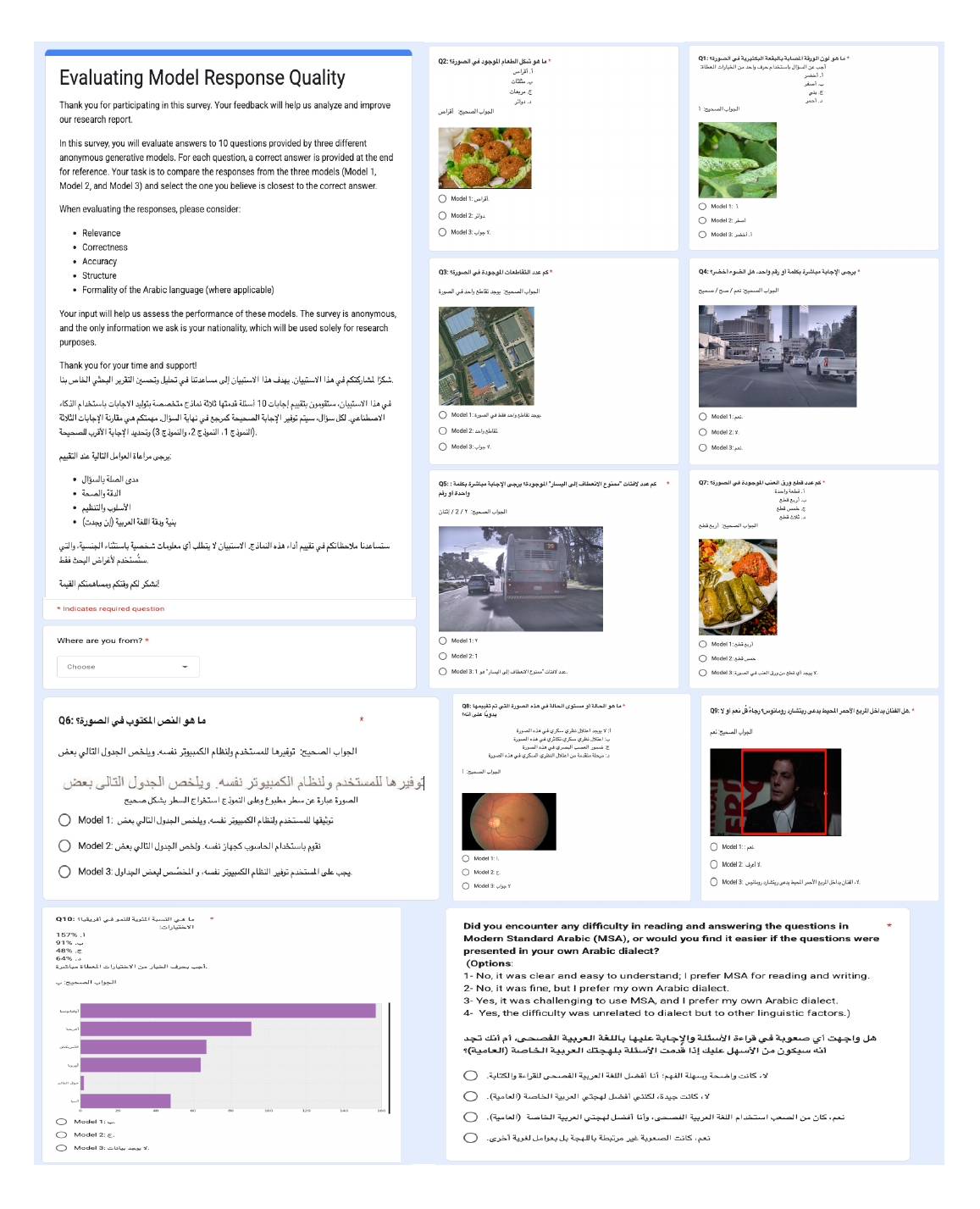}
  \caption{AIN human evaluation survey, illustrating assessment criteria and multi-domain questions designed to evaluate multi-task and complex reasoning capabilities. The survey includes evaluations on specific food items, road signs in low-resolution settings, celebrities, charts, remote sensing tasks, and other diverse topics to comprehensively assess performance across multiple domains and challenges.}
  \label{fig:ain_survey}
\end{figure}

For qualitative comparison, Figure~\ref{fig:ain_contre} highlights AIN's performance relative to open-source and closed-source LMMs (GPT-4o \cite{gpt4o} and LLaVA \cite{li2024llava}, respectively) across various domains. Unlike its counterparts, which frequently provided incorrect, incomplete answers or failed to adhere to the required format, AIN consistently delivers accurate and contextually appropriate responses. The model demonstrates proficiency in handling diverse query formats, effectively addressing both multiple-choice and open-ended questions with precision and reliability.

\subsection{Human Feedback}
To further evaluate our AIN model, we conduct a qualitative assessment through human feedback, comparing it against closed- and open-source LMMs in a blind setup, where model identities are not revealed to participants. The survey, as shown in Figure~\ref{fig:ain_survey}, covers various real-world domains, including medical diagnosis, road signs, and other scenarios such as low-resolution settings. Targeted at Arabic native speakers, it consists of 10 questions evaluating a range of topics, each with corresponding ground-truth answers. Participants are tasked with selecting the response from the three models that they believe are closest to the ground-truth.

In the survey, \textbf{``Model 1''} represented AIN-7B (ours), \textbf{``Model 2'' } corresponds to GPT-4o  \cite{gpt4o}, and \textbf{``Model 3''} is LLaVA \cite{li2024llava}. Delivered in MSA, the survey also includes an additional question to gather feedback on language clarity, MSA/ dialect preferences, and the survey itself.

\begin{figure}[hptb]
  \centering
  \includegraphics[width=0.85\textwidth]{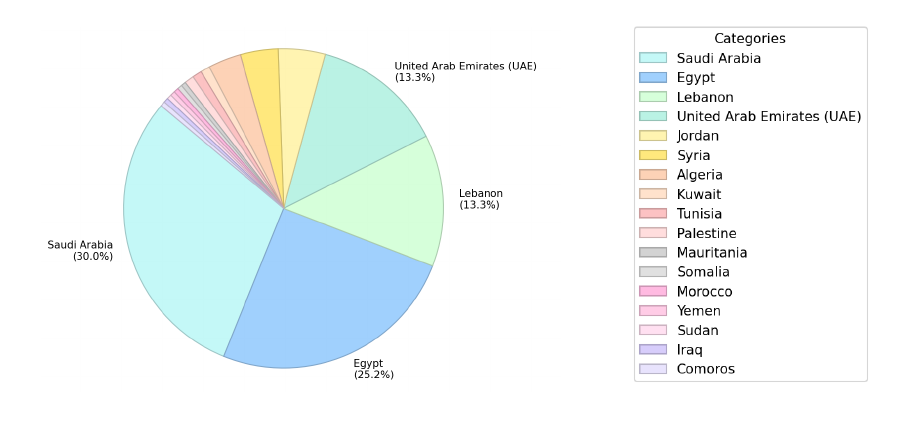}
  \caption{Nationality of AIN Survey Participants: Participants represent 17 Arab nations, with the highest contributions from Saudi Arabia (30\%), followed by Egypt (25\%), the UAE (13.3\%), and Lebanon (13.3\%).}
  \label{fig:ain_survey_nationality}
\end{figure}

\paragraph{Survey Participation.} More than 200 participants from 17 Arab countries (Figure~\ref{fig:ain_survey_nationality}), selected from diverse sectors and educational background, completed the survey. The highest contributions came from Saudi Arabia (30\%), followed by Egypt (25\%), the UAE (13.3\%), and Lebanon (13.3\%).

\begin{figure}[t]
  \centering
  \includegraphics[width=0.36\textwidth]{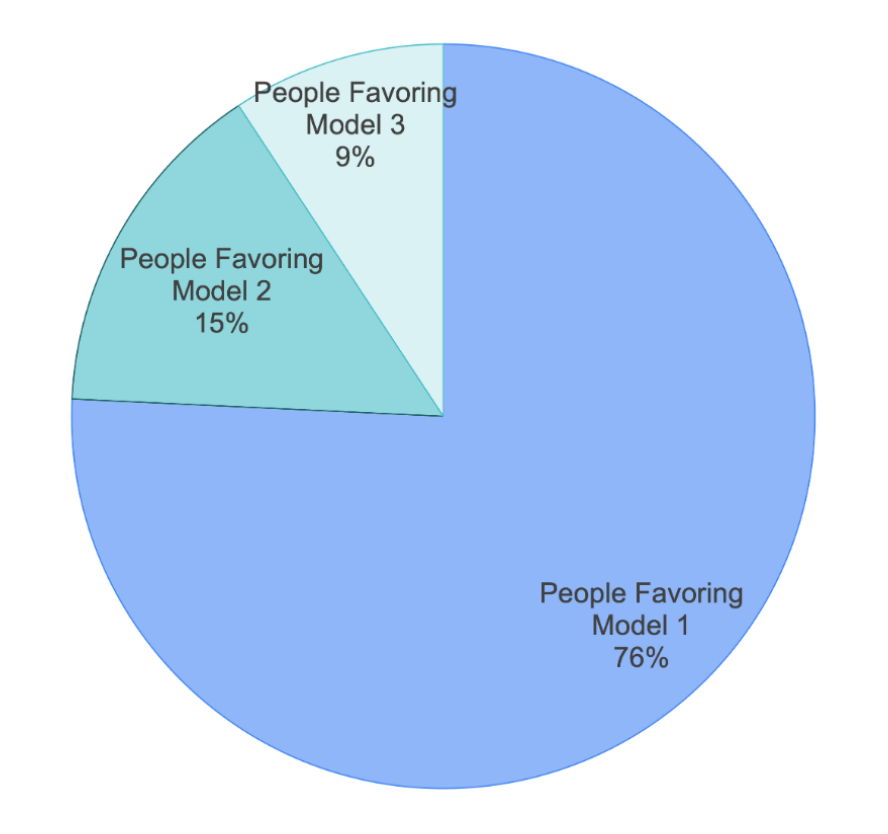}
  \caption{User Model Preferences. Participant preferences for the three models in the survey, with Model 1 (AIN (ours)) receiving 76\% of the votes, GPT-4o 15\%, and LLaVA 9\%, demonstrating AIN's strong performance.}
  \label{fig:ain_survey_favor}
  \label{fig:ain_survey_model_favor}
\end{figure}

\paragraph{Model Preferences.} Figure~\ref{fig:ain_survey_model_favor} shows that participants predominantly favor Model 1 (AIN-7B \emph{(ours)}), which received 76\% of the votes. GPT-4o followed with 15\%, and LLaVA garnered 9\%, underscoring AIN's significant preference among respondents.

\begin{figure}[hptb]
  \centering
  \includegraphics[width=\textwidth]{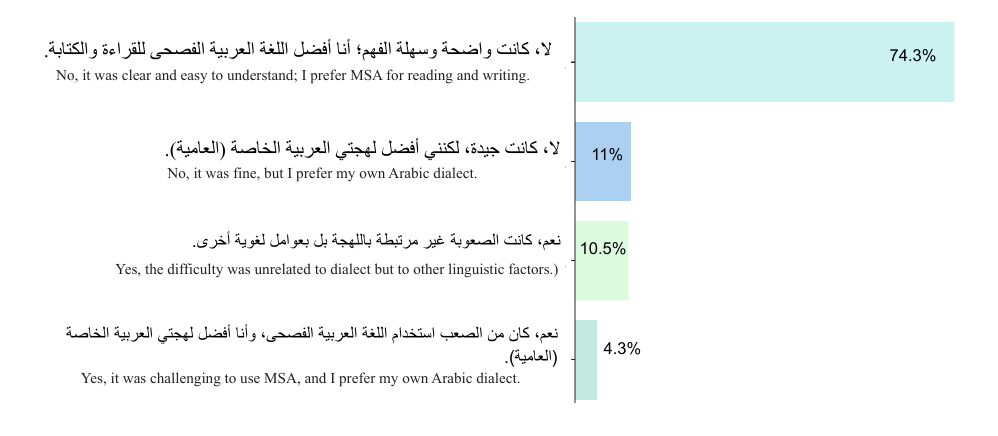}
  \caption{ User Preferences for MSA and Local Dialects: The majority (74.3\%) preferred MSA for reading and writing. An additional 11\% are comfortable with MSA but favored their local dialects, while 4.3\% found MSA challenging and preferred using their dialect. A further 10.5\% reported difficulties unrelated to linguistic aspects.}
  \label{fig:ain_survey_msa}
\end{figure}

\begin{figure}[hptb]
  \centering
  \begin{subfigure}[b]{0.45\textwidth} 
    \centering
    \includegraphics[width=\textwidth]{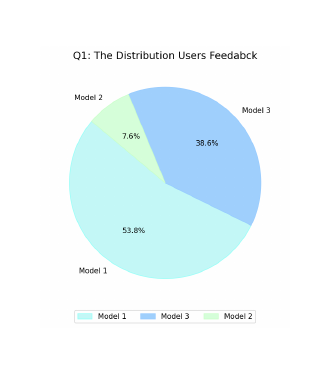}
    \caption{Q1:\scalebox{0.75}{ \< ما هو لون الورقة المصابة بالبقعة البكتيرية؟> }\\ \textbf{Domain: }	Agricultural Image Understanding / Plant diseases.\\ \textbf{Purpose: }Ability to detect diseased plant areas and identify their color.}
    \label{fig:survey_q1}
  \end{subfigure}
  \hfill 
  \begin{subfigure}[b]{0.45\textwidth} 
    \centering
    \includegraphics[width=\textwidth]{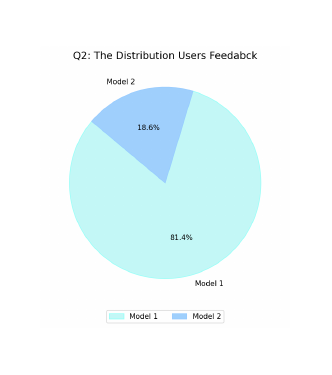}
    \caption{Q2:\scalebox{0.75}{\<ما هو شكل الطعام الموجود في الصورة؟> }\\ \textbf{Domain:} 	Cultural-Specific Image Understanding / Food.\\
\textbf{Purpose:} Ability to recognize food and precisely determine its shape.).}
    \label{fig:survey_q2}
  \end{subfigure}

  \vspace{1em} 
  \begin{subfigure}[b]{0.45\textwidth} 
    \centering
    \includegraphics[width=\textwidth]{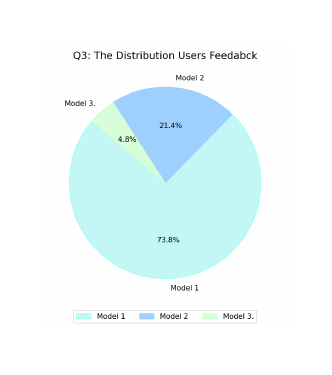}
    \caption{Q3:\scalebox{0.75}{\<كم عدد التقاطعات الموجودة في الصورة؟>} \\ \textbf{Domain: }	Remote Sensing Image Understanding / Roads \& Constructions. \\
\textbf{Purpose:} Ability to identify specific constructions among similar ones.}
    \label{fig:survey_q3}
  \end{subfigure}
  \hfill 
  \begin{subfigure}[b]{0.45\textwidth} 
    \centering
    \includegraphics[width=\textwidth]{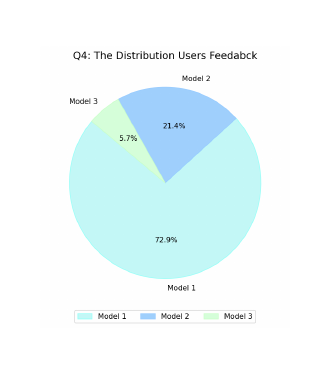}
    \caption{Q4:\scalebox{0.725}{ \<يرجى الإجابة مباشرة بكلمة أو رقم واحد، هل الضوء أخضر؟> }\\\textbf{ Domain:} 	General VQA/ Binary Question. \\
\textbf{Purpose:} Ability to identify tiny details in ambiguous scenes and answer binary questions. }
    \label{fig:survey_q4}
  \end{subfigure}
  \caption{Survey Feedback - Part 1: Questions 1 to 4 explored diverse domains, including agriculture, cultural-specific topics (e.g., food), remote sensing, and general VQA. Tasks included agro-disease detection, food recognition, shape identification, specific construction detection, and recognizing tiny details in ambiguous scenes, using various question formats such as MCQ, binary, and open-ended short answers.}
  \label{fig:survey_q1_4}
\end{figure}

\begin{figure}[hptb]
  \centering
  \begin{subfigure}[b]{0.45\textwidth} 
    \centering
    \includegraphics[width=\textwidth]{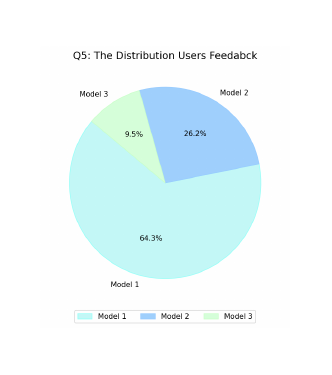}
    \caption{Q5:\scalebox{0.75}{ \<كم عدد لافتات ممنوع الانعطاف إلى اليسار الموجودة؟> }\\ \textbf{Domain: }General VQA / Traffic Signs.\\\textbf{Purpose:} Ability to spot traffic signs at a distance and in low resolution.}
    \label{fig:survey_q5}
  \end{subfigure}
  \hfill 
  \begin{subfigure}[b]{0.45\textwidth} 
    \centering
    \includegraphics[width=\textwidth]{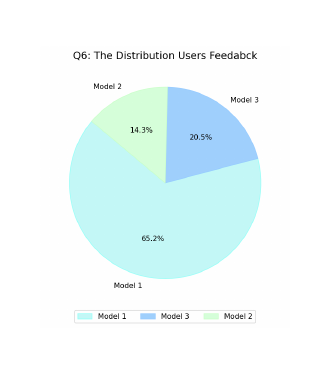}
    \caption{Q6:\scalebox{0.75}{\<ما هو النص المكتوب في الصورة؟ >} \\ \textbf{Domain:} OCR \& Document Understanding.\\
    \textbf{Purpose:} Ability to discern Arabic characters and extract text from images.}
    \label{fig:survey_q6}
  \end{subfigure}

  \vspace{1em} 
  \begin{subfigure}[b]{0.43\textwidth} 
    \centering
    \includegraphics[width=1.08\textwidth]{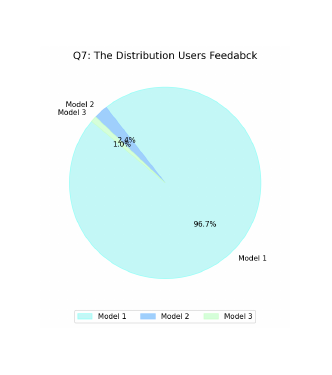}
    \caption{Q7:\scalebox{0.75}{\<كم عدد قطع ورق العنب الموجودة في الصورة؟>}\\ \textbf{Domain: }General VQA / Short Answer Question.\\
    \textbf{Purpose:} Ability to pinpoint the required item among several items + provide a short answer as required.}
    \label{fig:survey_q7}
  \end{subfigure}
  \hfill 
  \begin{subfigure}[b]{0.43\textwidth} 
    \centering
    \includegraphics[width=1.08\textwidth]{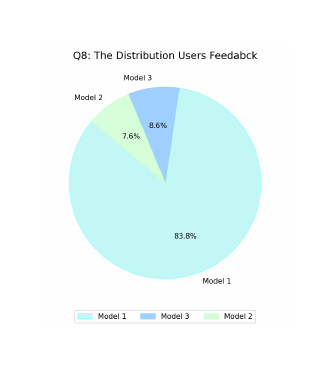}
    \caption{Q8:\scalebox{0.75}{ \<ما هو الحالة أو مستوى الحالة في هذه الصورة؟ >} \\ \textbf{Domain: }Medical Image Understanding / Diseases Diagnoses.\\
   \textbf{Purpose:} Ability to diagnose organ health by reasoning its condition (normal or abnormal) for a specific disease. }
    \label{fig:survey_q8}
  \end{subfigure}

  \caption{Survey Feedback - Part 2: Questions 5 to 8 focus on domains such as traffic sign recognition, OCR and document understanding, general VQA, and medical imaging. Tasks include identifying traffic signs, extracting correct text from images, pinpointing specific items among several options, and diagnosing organ conditions, using various formats such as MCQ and short answers.}
  \label{fig:survey_q5_8}
\end{figure}

\begin{figure}[hptb]
  \centering
  \begin{subfigure}[b]{0.39\textwidth} 
    \centering
    \includegraphics[width=6.75cm]{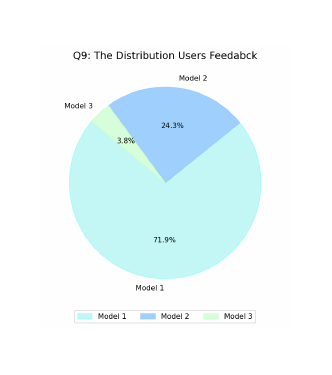}
    \caption{Q9:\scalebox{0.725}{ \<هل الفنان بداخل المربع الأحمر المحيط يدعى ريتشارد رومانوس؟> }\\
      \textbf{Domain: }General VQA / Grounding and Celebrities. \\
      \textbf{Purpose: }Ability to determine a person's identity in a specific location.}
    \label{fig:survey_q9}
  \end{subfigure}
  \hfill 
  \begin{subfigure}[b]{0.4\textwidth}
    \centering
    \includegraphics[width=6.75cm]{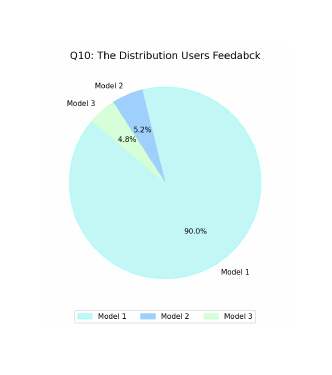}
    \caption{Q10: \scalebox{0.75}{\<ما هي النسبة المئوية للنمو في أفريقيا؟>} \\
      \textbf{Domain: }Chart, Diagram \& Table Understanding / Bar Charts. \\
      \textbf{Purpose: }Ability to extract values from charts, even when not explicitly shown.}
    \label{fig:survey_q10}
  \end{subfigure}

  \caption{Survey Feedback - Part 3: Questions 9 and 10 focus on domains such as celebrities, grounding, and charts and diagrams. Tasks include identifying a celebrity in a specific location within the image based on the question, and extrapolating values from charts where the information is not explicitly written, using formats such as MCQ and short answers.}
  \label{fig:survey_q9_10}
\end{figure}

\paragraph{Survey Results and Analysis.} The survey results reveal that AIN outperforms human participants in several questions, demonstrating notable advantages in accuracy and adherence to the response formats. For instance, in Q1, although both Model 3 and human participants provided correct answers, they failed to comply with the required response format, a criterion that AIN successfully fulfilled. Similarly, in Q9, 24.9\% of participants, and in Q6, 20.5\%, were unable to respond appropriately, whereas AIN provides accurate responses. In Q2, AIN exhibits superior precision by correctly identifying the shape as a disc rather than a circle, outperforming 18.6\% of human respondents. Furthermore, in Q3 and Q5, participants struggle to recognize small details in the images, highlighting AIN’s ability to detect subtle features. Notably, in Q10, AIN demonstrates its capacity to solve complex reasoning tasks by extracting value beyond the visible content of the image. This highlights its ability to handle abstract problem-solving, further showcasing its comparative advantage over other models. Detailed results and comparative analyses are illustrated in Figures~\ref{fig:survey_q1_4}, \ref{fig:survey_q5_8}, and \ref{fig:survey_q9_10}.

\paragraph{Dialect Preferences.} Regarding language suitability, depicted in Figure~\ref{fig:ain_survey_msa}, 74.3\% of participants found MSA appropriate for the survey. An additional 11\% were comfortable with MSA but expressed a preference for their local dialect. Only 4.3\% strongly preferred their local dialect over MSA, while 10.5\% reported challenges unrelated to language.

\newpage
\section{Data Inspection and Selection}
Our data collection comprises publicly available MSA Arabic and English datasets, with a portion specifically curated for model training. Notably, 35\% of the Arabic data is authentic. To achieve scalability and address specific domain requirements, translation was utilized to complement the existing data.

\subsection{Data Translation}
In pursuit of optimal data translation, we select three models from the GPT-4v suite—specifically, GPT-4 \cite{gpt4}, GPT-4o \cite{gpt4o}, and GPT-4o-mini \cite{gpt4o-mini}. A comprehensive experiment is conducted to evaluate the performance of these models across key criteria, including translation correctness, translation accuracy, and translation efficiency. For this evaluation, a variety of random samples of original English content are selected, and an English prompt was meticulously curated. To ensure accuracy and cultural adequacy, the corresponding Arabic prompts are crafted by a native Arabic speaker. Additionally, the same samples are manually translated by native Arabic speakers with high proficiency, serving as reference translations for benchmarking.

We evaluate the translation performance of GPT-4, GPT-4o, and GPT-4o-mini using both Arabic and English prompts under identical settings. \\
Translation accuracy is assessed by native speakers who rated the outputs against manually translated references on a scale from 0 (fail) to 1 (accurate). The results indicate that GPT-4o and GPT-4o-mini perform closely in speed and accuracy across both prompt types, with the Arabic prompt outperforming in both metrics. Notably, GPT-4o-mini achieves the highest accuracy with the Arabic prompt. Contributors generally found that GPT-4o-mini successfully translates all terms, including brand names such as ``Boeing'', whereas GPT-4o frequently fails to provide complete or accurate translations for these sentences.

Based on these findings, GPT-4o-mini is selected for data translation.
All reading and results are recorded in Table~\ref{tab:arabic_prompt_tab} and Table~\ref{tab:english_prompt_tab} for Arabic and English prompts, respectively.

\begin{table}[htpb]
\centering
\caption{Comparison of translation performance for Arabic prompts across three GPT-4v models, evaluated on a variety of samples. Avg time/iteration\textsuperscript{*}: average time per sample.}
\vspace{0.5em}
\renewcommand{\arraystretch}{1.5}
\setlength{\tabcolsep}{22pt} 
\resizebox{\textwidth}{!}{
\begin{tabular}{lccc}
\toprule
\rowcolor{gray!15}\multicolumn{4}{c}{\textbf{Arabic Prompt}} \\
\rowcolor{gray!10}
\textbf{Model}& \textbf{Time} & \textbf{Avg time/ iteration\textsuperscript{*}} & \textbf{Accuracy} \\
\hline
\textbf{GPT-4o \cite{gpt4o}} & 1 min, 43 sec & \textbf{04.16 sec} &90\%\\
\textbf{GPT-4 \cite{gpt4}} & 6 min, 39 sec & 15.99 sec & 85\%\\
\textbf{GPT-4o-mini \cite{gpt4o-mini}} & 1 min, 28 sec & 03.52 sec &\textbf{ 92\%}\\
\bottomrule
\end{tabular}}
\label{tab:arabic_prompt_tab}
\end{table}

\begin{table}[htpb]
\centering
\caption{Comparison of translation performance for English prompts across three GPT-4v models, evaluated on variety of samples. Avg time/iteration\textsuperscript{*}: average time per sample.} 
\vspace{0.5em}
\renewcommand{\arraystretch}{1.5}
\setlength{\tabcolsep}{22pt} 
\resizebox{\textwidth}{!}{
\begin{tabular}{lccc}
\toprule
\rowcolor{gray!15}\multicolumn{4}{c}{\textbf{ English Prompt}} \\
\rowcolor{gray!10}
\textbf{Model}& \textbf{Time} & \textbf{Avg time/ iteration \textsuperscript{*}} & \textbf{Accuracy} \\
\hline
\textbf{GPT-4o \cite{gpt4o}} & 2 min, 01 sec & 04.87 sec &\textbf{88\%}\\
\textbf{GPT-4 \cite{gpt4}} & 8min, 21 sec & 20.08 sec & 50\%\\
\textbf{GPT-4o-mini \cite{gpt4o-mini}} & 1 min, 52 sec &\textbf{ 4.48 sec} & 87\%\\
\bottomrule
\end{tabular}}
\label{tab:english_prompt_tab}
\end{table}

\subsection{Data Quality Verification and Filtering}
\begin{figure}[hptb]
    \centering
    \includegraphics[width=\linewidth,height=7.5cm]{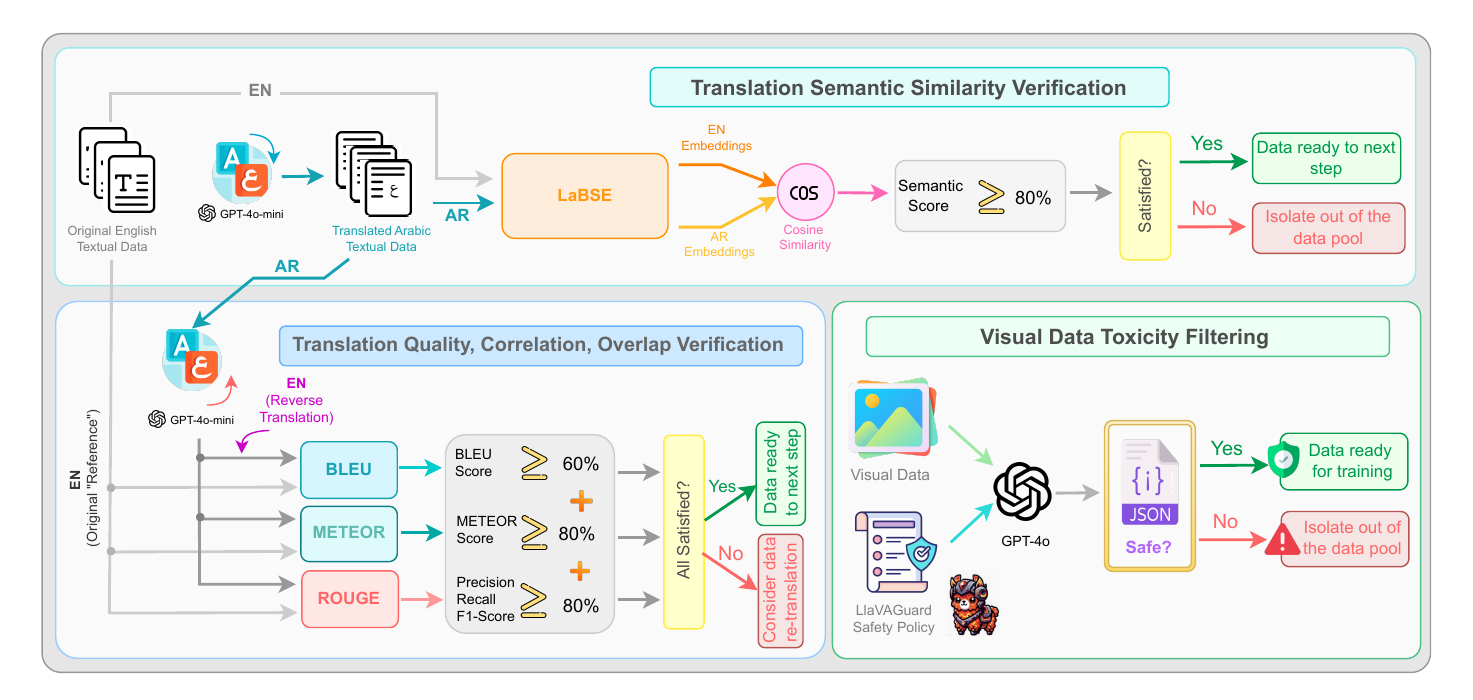}
    \caption{\textbf{Data verification and filtering pipeline for textual and visual data.} Textual data underwent semantic similarity checks using LaBSE \cite{feng2020language} (80\% threshold) and quality evaluation using BLEU \cite{papineni2002bleu} (60\% threshold), METEOR \cite{banerjee2005meteor} (80\% threshold), and ROUGE \cite{lin2004rouge} (80\% threshold). Visual data was screened for toxicity using LLavaGuard \cite{helff2024llavaguard} policies with GPT-4o \cite{gpt4o}, discarding unsafe images to ensure quality and safety.}
    \label{fig:qual_pipe}
\end{figure}

High-performance models inherently require high-quality data \cite{deitke2024molmo}. Therefore, in addition to selecting appropriate data, we have implemented a multi-step data translation verification procedure alongside rigorous toxicity-free filtering (Figure~\ref{fig:qual_pipe}).

\paragraph{Data Semantic Translation Verification.}
To identify the optimal model for translation verification, we design a set of sentences reflecting common linguistic challenges in Arabic, including punctuation alignment with English, direct and semantic translation accuracy, masculine/feminine tone differentiation, and handling of diacritics. This evaluation involves 21 sentence pairs, categorized as simple sentences (Table~\ref{tab:simple_sentence}), complex sentences with tone ambiguity (Table~\ref{tab:complex_sentence}), and affirmative clauses with question (Table~\ref{tab:QA_sentence}).

The sentence pairs are processed using five multilingual models—M-BERT \cite{kenton2019bert}, Paraphrase-XLM-R \cite{reimers-2019-sentence-bert}, all-mpnet-base-v2 \cite{song2020mpnet}, LaBSE \cite{feng2020language}, and AraBERT \cite{antoun2020arabert}—to evaluate semantic similarity between English and Arabic translations. Cosine similarity is used as the scoring metric to quantify the alignment between the translations, providing a robust basis for selecting the most suitable model.\\

\newcolumntype{R}[1]{>{\raggedleft\arraybackslash}p{#1}}

\begin{table}[htpb]
\centering
\caption{Translation quality check - Sentence 1: A simple English sentence with different settings including accurate direct translation, semantic translation, mismatched translation, punctuation, and diacritics.}
\vspace{0.5em}
\setlength{\tabcolsep}{9pt} 
\resizebox{\textwidth}{!}{
\begin{tabular}{p{0.5cm}p{4cm}R{3cm}p{6cm}}
\hline
\rowcolor{gray!10}
\textbf{Ref.}&\textbf{Original} & \textbf{Translation} & \textbf{Translation Criteria} \\
\hline
1.1&This is an example sentence & \<هذه جملة مثال> & Accurate direct translation \\
1.2&This is an example sentence & \<مِن فَضلِكِ اِجلِس >& Completely mismatched translation. \\
1.3&This is an example sentence & \<هذه عبارة توضيحية > &Translation with semantic meaning.\\
1.4&This is an example sentence & \<هَذِهِ جُمْلَةٌ مِثَالٌ> & Accurate direct translation + diacritics.\\
1.5&This is an example sentence! & \<هذه عبارة توضيحية!> & Translation with semantic meaning + punctuation, no diacritics \\
\hline
\end{tabular}}
\label{tab:simple_sentence}
\end{table}

\begin{table}[htpb]
\centering
\caption{Translation quality check - Sentence 2: The English sentence consists of a polite request with different settings including accurate direct translation, semantic translation, masculine/ Feminine tone, mismatched translation, punctuation, and diacritics.}
\vspace{0.5em}
\setlength{\tabcolsep}{9pt} 
\resizebox{\textwidth}{!}{
\begin{tabular}{m{0.5cm}m{2.75cm}R{3cm}m{7.5cm}}
\hline
\rowcolor{gray!10}\textbf{Ref.}&\textbf{Original} & \textbf{Translation} & \textbf{Translation Criteria} \\
\hline
2.1&Please, sit down & \<تفضل، اجلس> & Accurate direct translation \\
2.2&Please sit down & \<تفضل اجلس> & No punctuation, no diacritics.\\
2.3&Please sit down & \<مِن فَضلِكِ اِجلِس> & No punctuation, with diacritics \\
2.4&Please sit down & \<تَفَضَّلي اِجلِسي> & No punctuation, with diacritics, feminine tone \\
2.5&Please sit down & \<تَفَضَّل اِجلِس> & No punctuation, with diacritics, masculine tone \\
2.6&Please, sit down & \<هذه عبارة توضيحية > & Completely mismatched translation.\\
2.7&Please, sit down & \<مِن فَضلِكِ اِجلِس> &Translation with semantic meaning.\\
2.8&Please, sit down & \<تَفَضَّل اِجلِس> & Punctuation + diacritics/ masculine tone. \\
 2.9&Please, sit down & \<تَفَضَّل اِجلِسي> & Accurate direct translation + diacritics/ partial feminine tone.\\
2.10&Please, sit down & \<تَفَضَّلي اِجلِسي>& Accurate direct translation + diacritics/ feminine tone.\\
2.11&Please, sit down. & \<مِن فَضلِكِ، اِجلِس.> & Accurate direct translation + punctuation + diacritics. \\
2.12&Please, sit down. & \<من فضلك، اجلس.> & Accurate direct translation + punctuation, no diacritics.\\
2.13&Please, sit down. & \<من فضلك، إجلس. >& With punctuation and only ``hamzat al kaser'' \<(إ)> \\
\hline
\end{tabular}}
\label{tab:complex_sentence}
\end{table}

\begin{table}[hptb]
\centering
\caption{Translation quality check - Sentence 3: The English sentence consists of an affirmative clause followed by a question of accurate direct translation in different settings including semantic translation and punctuation.}
\vspace{0.5em}
\setlength{\tabcolsep}{9pt} 
\resizebox{\textwidth}{!}{
\begin{tabular}{m{0.5cm}m{3.75cm}R{6.5cm}m{3.25cm}}
\hline
\rowcolor{gray!10}\textbf{Ref.}&\textbf{Original} & \textbf{Translation } & \textbf{Translation Criteria} \\
\hline
3.1&It is raining today should we stay at home & \<إنها تمطر اليوم هل يجب علينا البقاء في المنزل> & No punctuation. \\
3.2&It is raining today. Should we stay at home & \<إنه يوم ممطر هل يجب علينا البقاء في المنزل؟> & Semantic meaning + punctuation. \\
3.3&It is raining today. Should we stay at home? & \<إنها تمطر اليوم. هل يجب علينا البقاء في المنزل؟> & With punctuation. \\
\hline
\end{tabular}}
\label{tab:QA_sentence}
\end{table}

\begin{figure}[htpb]
  \centering
  \begin{subfigure}[b]{1\textwidth} 
    \centering
    \includegraphics[width=1.02\textwidth]{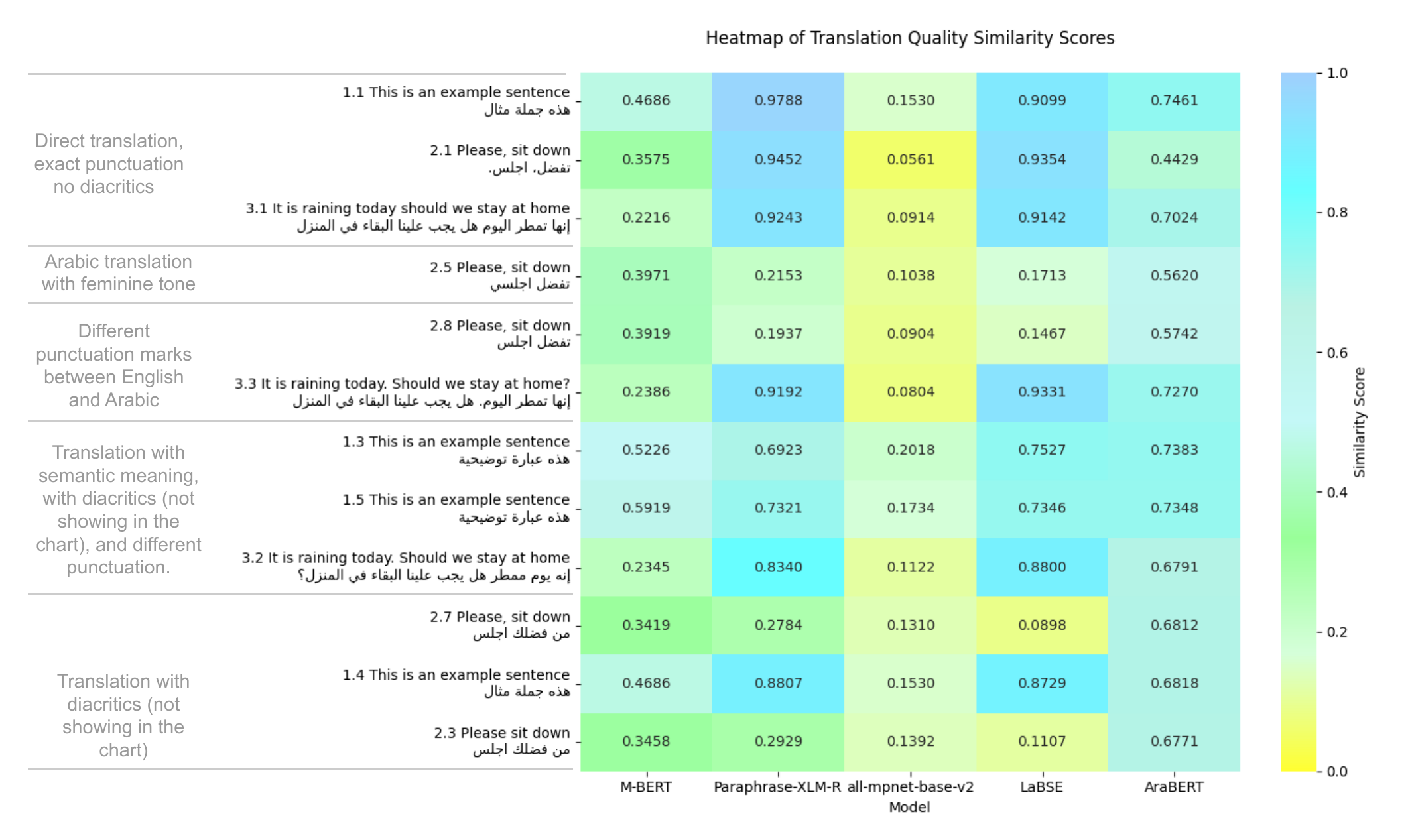}
    \caption{Similarity scores for different settings for correct translation. The higher the better.}
    \label{fig:heat_correct}
  \end{subfigure}
  \begin{subfigure}[b]{1\textwidth}
    \centering
    \includegraphics[width=1.02\textwidth]{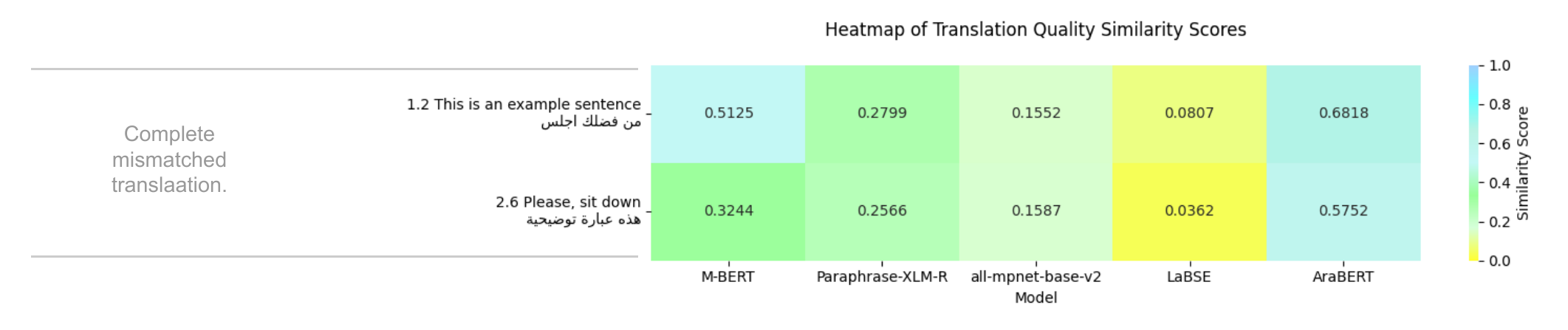} 
    \caption{Similarity scores for different settings for incorrect translation. The lower the worse.}
    \label{fig:heat_incorrect}
  \end{subfigure}
  \caption{Similarity scores for diverse settings, including direct correct translation, incorrect translation, semantic translation, diacritics, and punctuation.}  
  \label{fig:heatmap}
\end{figure}

Figure~\ref{fig:heat_correct} presents a heatmap of the similarity scores of the models in evaluating translations semantic correctness, considering punctuation, tone, and diacritics. While high similarity scores indicate good performance, a robust model must also assign low scores to poor or irrelevant translations. To assess this, a second experiment tested model behavior on mismatched translations ( Figure~\ref{fig:heat_incorrect}).

\begin{figure}[htpb]
  \centering
  \begin{subfigure}[b]{1\textwidth} 
    \centering
    \includegraphics[width=1.05\textwidth]{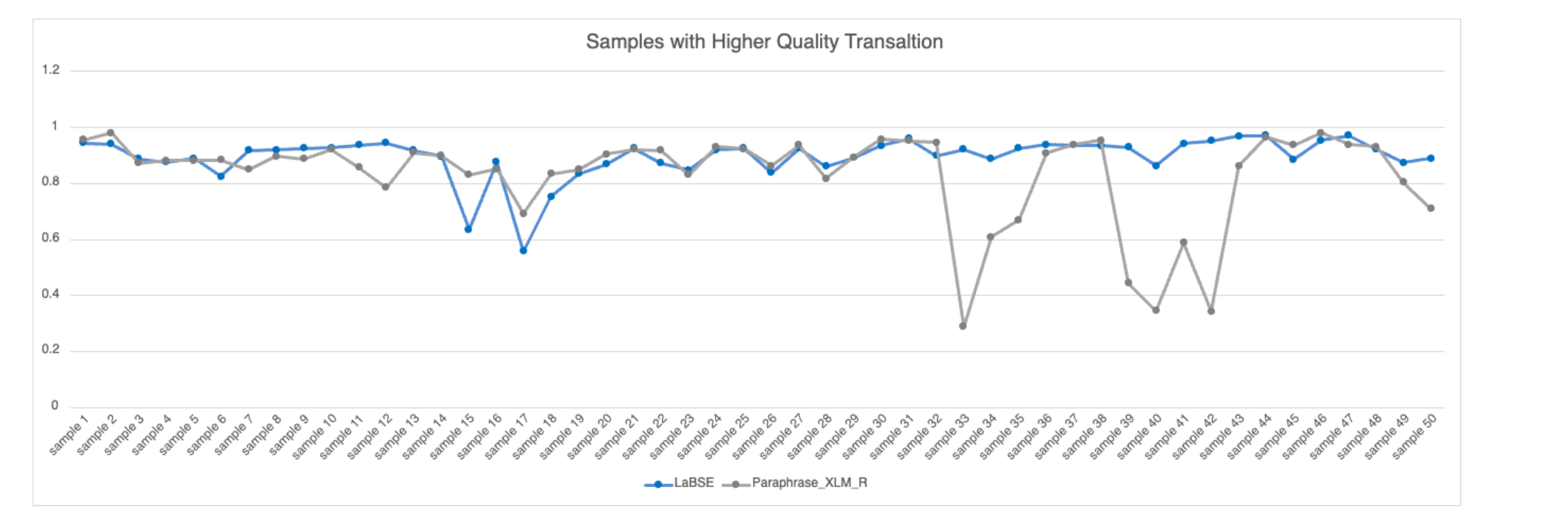}
    \caption{Comparison of LaBSE and Paraphrase-XLM-R: Evaluating 50 high-quality translated samples.}
    \label{fig:labse_vs_para_best}
  \end{subfigure}
  \begin{subfigure}[b]{1\textwidth}
    \centering
    \includegraphics[width=\textwidth]{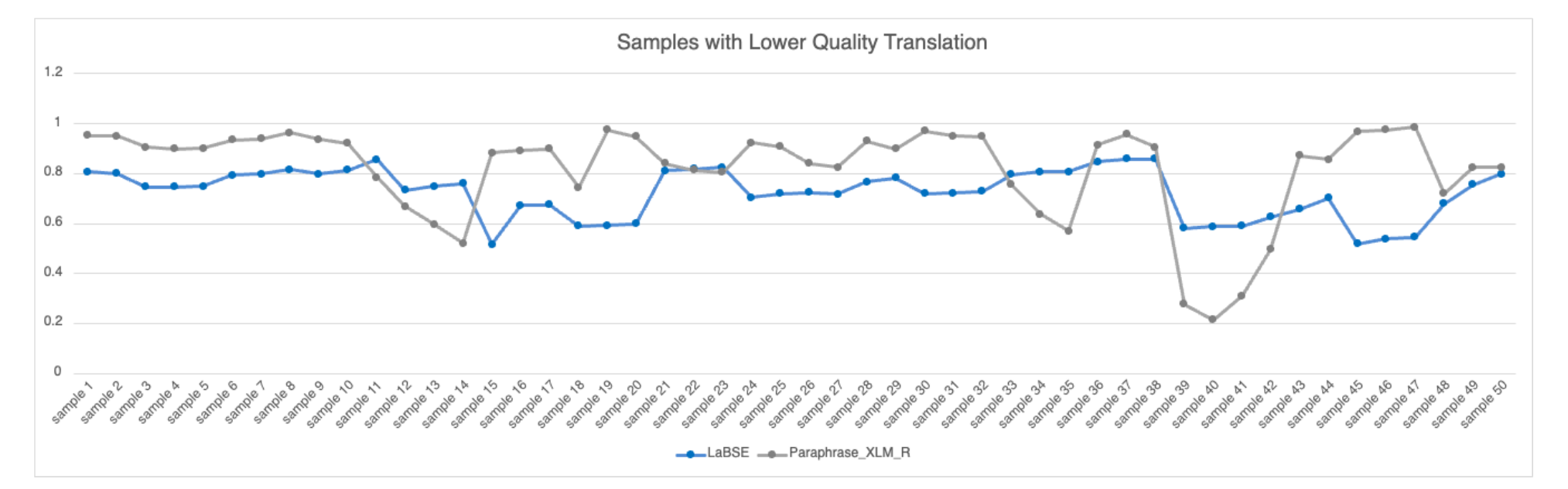} 
    \caption{Comparison of LaBSE and Paraphrase-XLM-R: Assessing 50 low-quality translated samples.}
    \label{fig:labse_vs_para_poor}
  \end{subfigure}
  \caption{Comparison of LaBSE and Paraphrase-XLM-R to identify the optimal model.}  
  \label{fig:labse_vs_para_low}
\end{figure}

Due to the close performance observed between LaBSE \cite{feng2020language} and Paraphrase-XLM-R \cite{reimers-2019-sentence-bert} in initial evaluations, an additional experiment is conducted to further assess their capabilities. This experiment utilized 50 samples of high-quality translations and 50 samples of moderate-to-poor translations. LaBSE demonstrated superior consistency, providing higher similarity scores for accurate translations and relatively lower scores for poor translations compared to Paraphrase-XLM-R. This reliability in distinguishing translation quality (Figures~\ref{fig:labse_vs_para_best} and \ref{fig:labse_vs_para_poor}) led to the selection of LaBSE for the full dataset. Translations scoring below 80\% similarity were excluded, accounting for less than 2\% of the data.

\paragraph{Data Quality Verification.}

To ensure a comprehensive quality verification process, particularly given the use of generative AI for translation, additional checks are necessary. These include assessing the quality of machine-generated text, its correlation with the original content, and the degree of overlap between the generated and original text. To achieve this, we employ three specialized evaluation metrics: BLEU (2-gram and 4-gram) \cite{papineni2002bleu} for text quality, METEOR \cite{banerjee2005meteor} for translation correlation, and ROUGE (unigram and ROUGE-L) \cite{lin2004rouge} for overlap measurement.

Our data quality verification experiment involves analyzing randomly selected 50 translated samples that are translated back to English using GPT-4o-mini \cite{gpt4o-mini} for comparison against the original English text (reference data). 

The evaluation metrics demonstrated strong performance across multiple dimensions: BLEU scores \cite{papineni2002bleu} of 71.11\% (2-gram) and 60.20\% (4-gram) indicated high local coherence and good fluency in the translated text. The METEOR score \cite{banerjee2005meteor} of 86.10\% suggested high-quality translation with effective handling of both exact matches and linguistic variations. ROUGE metrics \cite{lin2004rouge} were particularly strong, with unigram scores showing 87.80\% precision and 87.30\% recall, indicating excellent word-level accuracy and comprehensive content capture. Similarly, ROUGE-L scores \cite{lin2004rouge} (precision: 86.20\%, recall: 85.90\%, F1: 85.80\%) confirmed strong structural similarity between the translated and reference texts, demonstrating that the essential meaning and structure are well-preserved throughout the translation process (Table~\ref{tab:evaluation_metrics}).

With the data translation quality checks completed, a final step of visual toxicity inspection is required to ensure the data is ready for model training.

\begin{table}[htpb]
\centering
\caption{Data Quality Verification and Evaluation Metrics}
\renewcommand{\arraystretch}{1.5}
\setlength{\tabcolsep}{20pt} 
\resizebox{\textwidth}{!}{
\begin{tabular}{lccccl}
\hline
\rowcolor{gray!10}\textbf{Metric} & \textbf{Scores} & \textbf{Precision} & \textbf{Recall} & \textbf{F1-score} \\
\hline
BLEU (2-gram) \cite{papineni2002bleu} & 71.11\% & \cellcolor{gray!15} & \cellcolor{gray!15} & \cellcolor{gray!15} \\
BLEU (4-gram) \cite{papineni2002bleu} & 60.20\% & \cellcolor{gray!15} &\cellcolor{gray!15} & \cellcolor{gray!15} \\
METEOR \cite{banerjee2005meteor}& 86.10\% & \cellcolor{gray!15} & \cellcolor{gray!15} & \cellcolor{gray!15}\\
\hline
ROUGE (unigram) \cite{lin2004rouge}& \cellcolor{gray!15}&87.80\% & 87.30\% &87.30\%\\
ROUGE-L \cite{lin2004rouge} & \cellcolor{gray!15}& 86.20\% &85.90\%& 85.80\% \\
\hline
\end{tabular}
}
\label{tab:evaluation_metrics}
\end{table}

\paragraph{Toxicity Filtering:}
\begin{figure}[hptb]
  \centering
  \includegraphics[width=.5\textwidth]{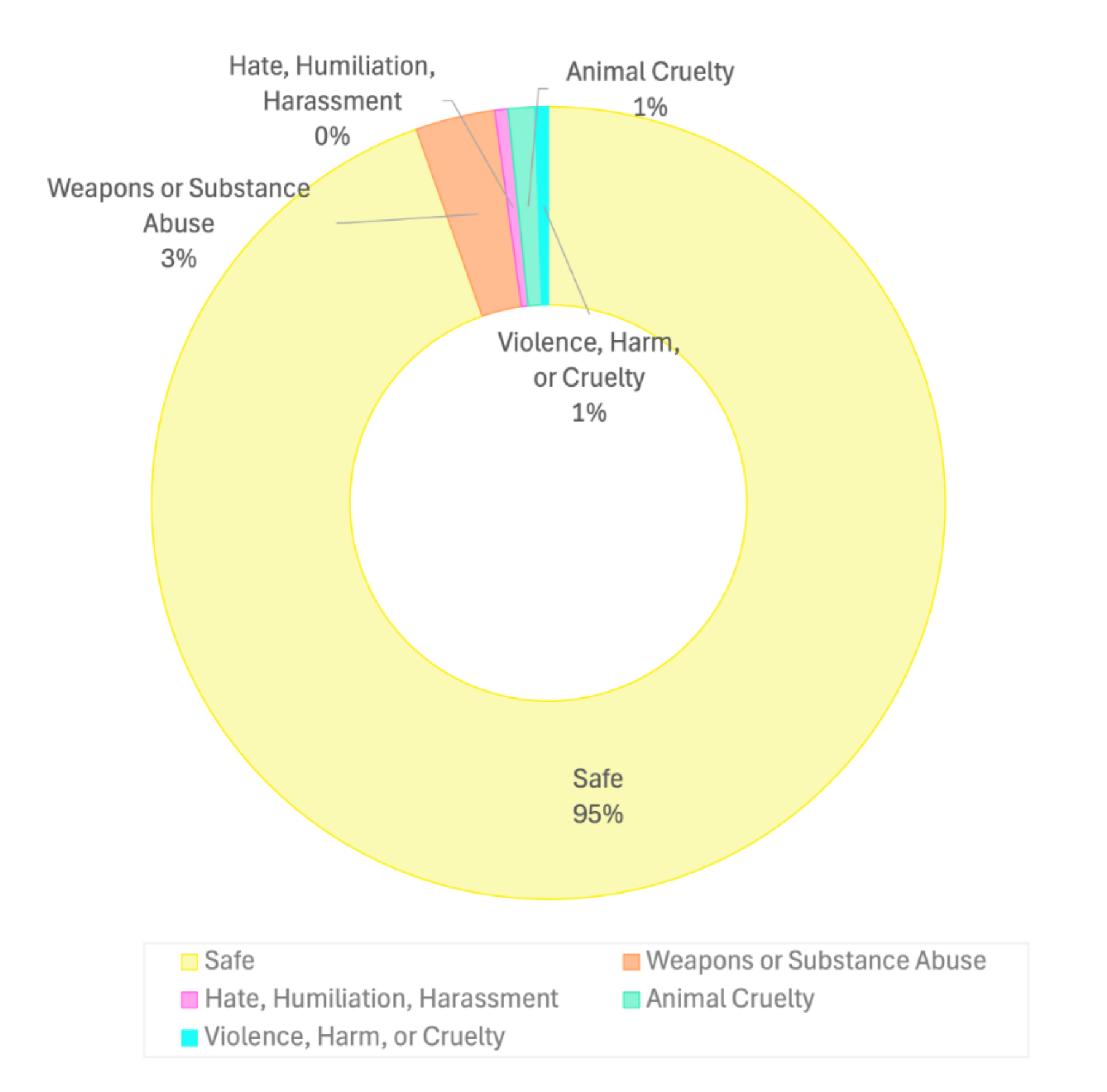}
  \caption{ Visual Data Toxicity Filtering. Using GPT-4o \cite{gpt4o} and LLavaGuard \cite{helff2024llavaguard} policies, about 96\% of the data is classified as safe, while the remainder was deemed unsafe. The unsafe data was distributed across four categories: ``Weapon, or Substance Abuse'' (3.25\%), ``Hate, Humiliation, Harassment'' (0.55\%), ``Animal Cruelty'' (1.09\%), and ``Violence, Harm, or Cruelty'' (0.55\%). }
  \label{fig:toxicity_distribution}
\end{figure}

To ensure model safety in vision, toxicity inspection is a critical component of our evaluation process. We utilize LLavaGuard's safety taxonomy \cite{helff2024llavaguard}, a well-curated prompt specifically designed to verify visual data against predefined safety criteria, in combination with GPT-4o for the inspection process. The dataset undergoes a comprehensive assessment to ensure compliance with safety policies. The evaluation covers key categories, including ``Hate, Humiliation, or Harassment''; ``Violence, Harm, or Cruelty''; ``Sexual Content''; ``Nudity''; ``Weapons or Substance Abuse''; ``Self-Harm''; ``Animal Cruelty''; and ``Disasters or Emergencies''. 

The results reveal that 95.63\% of the data is deemed safe, while 4.37\% is classified as unsafe. The unsafe data is distributed across four main categories: ``Weapons or Substance Abuse'', ``Hate, Humiliation, or Harassment'', ``Animal Cruelty'', and ``Violence, Harm, or Cruelty'', as illustrated in Figure~\ref{fig:toxicity_distribution}. This rigorous evaluation ensured that our data met safety standards, further preparing it for downstream applications.

Following the completion of the data verification steps and toxicity filtering, our dataset, comprising 3.6 million safe and curated entries, is prepared for model training.

\section{Experiments}
\label{experiment}

The AIN model is trained on 8 GPU nodes, each equipped with 8 NVIDIA A100 GPU cards, each with 80 GB memory. The GPUs within each node are interconnected using 8 NVLink links, ensuring high bandwidth and low latency. To facilitate efficient cross-node communication, each node is equipped with dual-port 200 Gbps (4×HDR) InfiniBand connections, achieving an aggregate interconnect bandwidth of 800 Gbps. This robust infrastructure was crucial in handling the computational and memory-intensive large-scale LMM training.

Our approach leverages the Qwen2-VL-7B \cite{bai2023qwen} model as the base, which we fine-tuned on our English-Arabic bilingual dataset. The dataset comprises 3.6 million high-quality text samples curated from diverse sources, ensuring comprehensive coverage of linguistic, cultural, and domain-specific nuances. For fine-tuning, we employed a full-parameter fine-tuning strategy, conducting training for one epoch. This approach allowed us to adapt the pre-trained model to better capture the semantic properties of both Arabic and English languages.

To optimize training efficiency and scalability, we employ the flash-attention mechanism, which significantly reduces memory overhead during attention computation. Additionally, we adhere to the hyper-parameter configurations established by LLaMA-Factory~\cite{zheng2024llamafactory}, including an optimized learning rate schedule, batch size, and weight decay strategies tailored for large-scale transformer-based models.

\section{Conclusion}
This work introduces AIN, an Arabic-inclusive LMM, as a step toward bridging the gap in AI solutions for Arabic, a low-resource yet globally significant language.  AIN is trained on a large-scale bilingual dataset with 35\% of authentic Arabic data. Through rigorous evaluation, we show that AIN achieves state-of-the-art performance across a wide range of tasks, including VQA, OCR and document understanding, cultural understanding, and domain-specific applications such as medical imaging and remote sensing, surpassing even bigger and more sophisticated models.
AIN further demonstrates superior accuracy, contextual understanding, and human-like reasoning in MSA, as validated by extensive evaluations and human judgments. By integrating advanced data curation, robust translation pipelines, and stringent quality control, AIN provides a new state-of-the-art multimodal AI model tailored to Arabic speakers.

\newpage
\bibliographystyle{unsrt}
\bibliography{ref}

\end{document}